\documentclass{article}
\usepackage[a4paper, total={6in, 9.5in}]{geometry}
\usepackage[dvipsnames]{xcolor}
\usepackage{mathtools} \usepackage{multirow}
\usepackage{booktabs}
\usepackage{colortbl}
\usepackage{subfigure}
\usepackage[export]{adjustbox}
\usepackage{afterpage}
\usepackage{bm}
\usepackage{wrapfig}
\usepackage{amsmath}
\usepackage{caption}

\usepackage[utf8]{inputenc} 
\usepackage[T1]{fontenc}    
\usepackage{hyperref}       
\usepackage{url}            
\usepackage{booktabs}       
\usepackage{amsfonts}       
\usepackage{nicefrac}       
\usepackage{microtype}      
\usepackage{xcolor}         

\def\eg{\emph{e.g.~}}

\def\ie{\emph{i.e.~}}
\def\aka{\emph{a.k.a.~}}

\DeclareMathOperator{\sign}{sign}

\usepackage{tikz}

\title{A New Kind of Adversarial Example}

\author{
Ali Borji \\
Quintic AI, San Francisco, CA \\ 
\texttt{aliborji@gmail.com} 
}

\begin{document}

\maketitle

\begin{abstract}
Almost all adversarial attacks are formulated to add an imperceptible perturbation to an image in order to fool a model. Here, we consider the opposite which is adversarial examples that can fool a human but not a model. A large enough and perceptible perturbation is added to an image such that a model maintains its original decision, whereas a human will most likely make a mistake if forced to decide (or opt not to decide at all). Existing targeted attacks can be reformulated to synthesize such adversarial examples. Our proposed attack, dubbed NKE, is similar in essence to the fooling images, but is more efficient since it uses gradient descent instead of evolutionary algorithms. It also offers a new and unified perspective into the problem of adversarial vulnerability. Experimental results over MNIST and CIFAR-10 datasets show that our attack is quite efficient in fooling deep neural networks. Code is available at \url{https://github.com/aliborji/NKE}.


\end{abstract}

\section{Introduction}

\begin{figure}[h]       
    \centering
    \includegraphics[width=.33\linewidth]{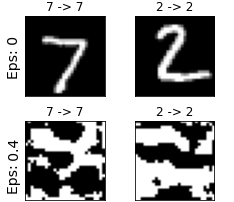}
    \includegraphics[width=.31\linewidth]{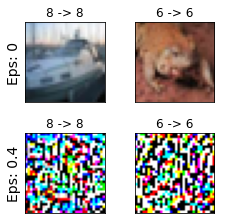}
    \caption{Sample adversarial examples generated by our attack for MNIST and CIFAR-10 images. The original images (top row) are highly perturbed (bottom row) such that the model still maintains its decision, whereas humans can no longer make a decision (or will make an error if forced to decide).}
    \label{fig:teaser}
\end{figure}

Deep neural networks are very brittle~\cite{szegedy2013intriguing,goodfellow2015explaining,borjiview2022}. A small imperceptible perturbation is enough to fool them. A sample $x'$ is said to be an adversarial example for $x$ when $x'$ is close to $x$ under a specific distance metric, while $f(x') \neq y$. Formally:
\begin{equation}
    \label{eq:def}
    x': D(x,x') < \epsilon_s \ \ \text{s.t} \ \ f(x') \neq y 
\end{equation}
where $D(.,.)$ is a distance metric, $\epsilon_s$ is a predefined distance constraint (\aka allowed perturbation), $f(.)$ is a neural network, and $y$ is the true label of sample $x$.

\begin{figure}[t]       
    \centering
    \includegraphics[width=.7\linewidth]{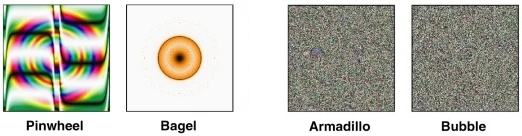}
        \caption{Example meaningless patterns (\aka fooling images) that are classified as familiar objects by a deep neural network~\cite{nguyen2015deep}.}
    \label{fig:fooling}
\end{figure}

The bulk of research on adversarial attack has focused on crafting adversarial examples by adding imperceptible perturbations to images. Some works have shown that perceptible perturbations can fool neural networks in the physical work (\eg~\cite{eykholt2018robust,sharif2019general,nguyen2015deep}). Here, we introduce a new attack which is in essence similar to the existing ones but differs in an important aspect. {\bf Instead of fooling a model to change its decision on the perturbed image (small imperceptible perturbation; images still recognizable by humans), it aims to fool the model to keep its original decision when there is almost no signal in the image (large perceptible perturbation; image unrecognizable by humans).} Our attack is similar to the fooling images~\cite{nguyen2015deep}, but instead of evolutionary techniques here we employ gradient descent which is much more efficient (See Fig.~\ref{fig:fooling}). In a sense, our work gives a new perspective to unify adversarial examples and fooling images under one umbrella. Gradient descent can be used to generate both types of perturbations to fool a model\footnote{Except images with patterns and textures similar to the images in the target class. See the left images in Fig.~\ref{fig:fooling}.}. Adversarial images synthesized by our attack for some images from MNIST and CIFAR-10 datasets are shown in Fig.~\ref{fig:teaser}. A model confidently maintains its original decision on the perturbed images that look just like noise to humans.



Highly perturbed images with the predicted label as the original label are dangerous and equally important as adversarial examples with imperceptible perturbations. As a real-world scenario, imagine you see a green light from a distance while driving. As you get closer to the light, it starts to rain such that you are no longer certain about the state of the light. In this situation, you may opt to slow down or pull over. Imagine a self-driving vehicle in the same situation, except now an adversary is manipulating the image of the light. It would be very dangerous if the car maintains its \emph{green} decision with high confidence, while the light may actually be red. Here, the adversary aims to maximize the chance of fooling the system by adding a highly perceptible perturbation.

\begin{figure}[hb]       
    \centering
    \includegraphics[width=.8\linewidth]{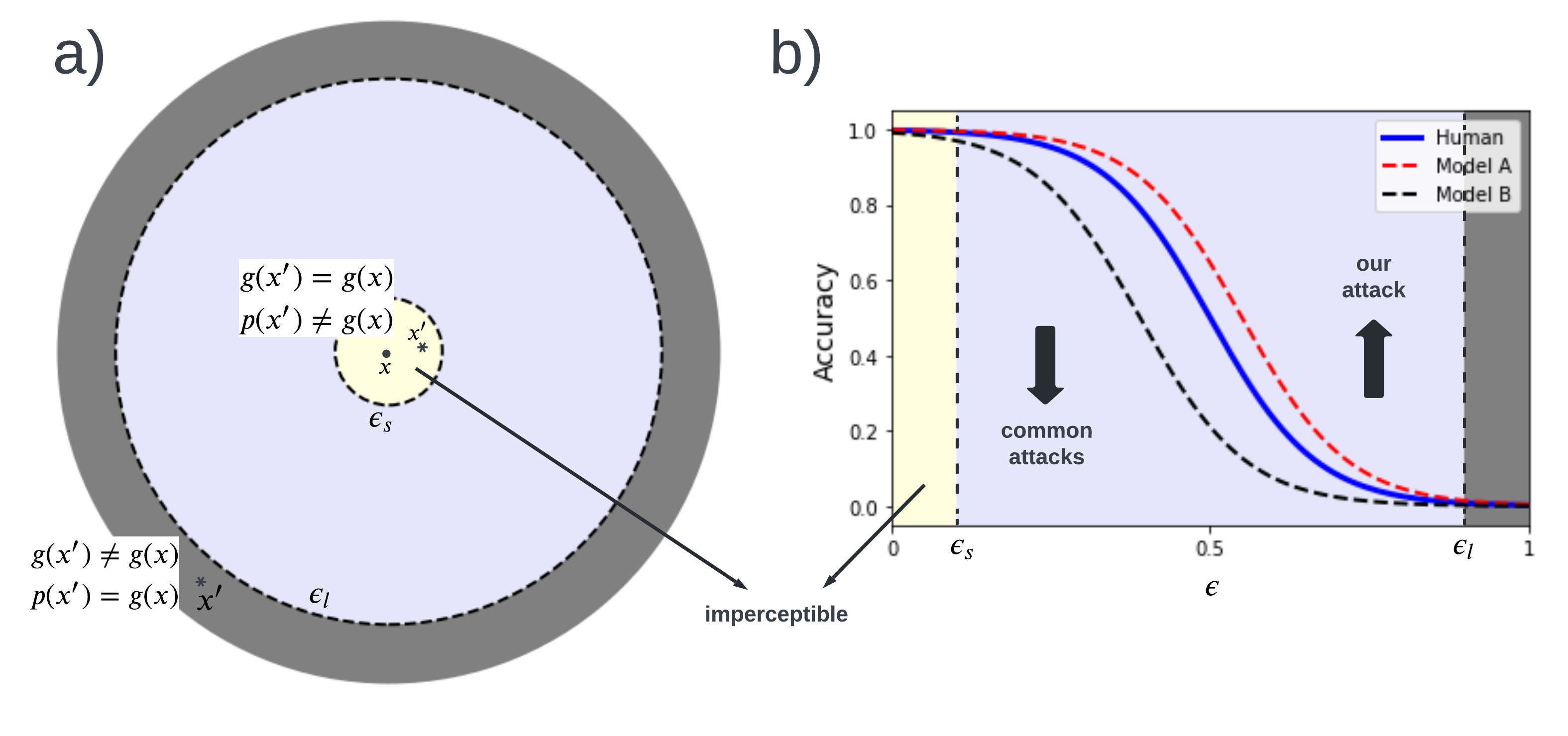}
        \caption{A general view of adversarial robustness. Hypothetically, a system is fully robust if it can function perfectly regardless of the perturbation magnitude. In reality, however, the performance curves look like a reverse sigmoid function (\aka psychometric functions for humans) indicating that performance drops as the perturbation grows (panel b). a) Current adversarial attacks are formulated to lower the left side of the curve (\ie zero accuracy for all values of $\epsilon$) and emphasize on imperceptible perturbations within the yellow $L_p$ ball (Eq.~\ref{eq:def}). In contrast, our attack aims to lift the right side of the curve (\ie 100\% accuracy for all values of $\epsilon$). These two types of attacks can also be combined, \ie lowering the left side and lifting the right side like a sigmoid. We emphasize on large perturbations (gray region) for which humans are more likely to fail. b) Hypothetical performance curves of humans vs. two models, one under performing humans (model A), and the other outperforming humans (model B). Here, model B is less robust than model A since its psychometric function falls below the psychometric function of model A.}
    \label{fig:newdef}
\end{figure}

\section{Attack formulation} 
We name our attack NKE after the ``new kind of adversarial example'' in the paper title. 
Adversarial examples generated by NKE fulfill the following condition:
\begin{equation}
    \label{eq:nke}
    x': \Bigl[ g(x') \neq g(x) \ \ \land  \ \ f(x') = g(x) \Bigr]
\end{equation}
where $g(x)$ is the oracle ($y$ in Eq.~\ref{eq:def}). Here, we assume that the initial prediction for sample $x$ is correct (\ie $f(x) = g(x)$). According to this formulation, an input is adversarial if the model predicts the same label as the original label while humans\footnote{human expert or the best possible classifier in a Bayesian sense.} will make a mistake if forced to make a decision (or it may opt not to make a decision at all). Since the function $g(x)$ is usually not available, a distance metric can be used as a proxy. We revise the Eq.~\ref{eq:def} to reflect the above definition as follows:
\begin{equation}
    \label{eq:def_revised}
    x': \Bigl[D(x,x') > \epsilon_l \ \ \land \ \ f(x') = y \Bigr]
\end{equation}
This new definition mirrors the classic definition. Instead of small imperceptible perturbations, it emphasizes highly perceptible perturbations (beyond a certain threshold $\epsilon_l$).  



Fig.\ref{fig:newdef} depicts a general view on adversarial robustness. A robust system should be insensitive to small perturbations while being sensitive to large perturbations. It should become more sensitive as the perturbation grows. Two perturbation bounds are particularly interesting. The first one, $\epsilon_s$, is considered in the classic formulation in Eq.~\ref{eq:def}. It is used to constrain the amount of the added perturbation (\ie \emph{maximum allowed perturbation} \eg 8/255)\footnote{The intuition here is that humans will not make a mistake at small perturbations, thus $g(x') = g(x)$.}. The second one, $\epsilon_l$, determines the \emph{minimum allowed perturbation}, and is chosen such that the perturbations beyond it will enforce a human error. The perturbed image would be considered adversarial if the model maintains its prior decision. The adversarial examples generated this way are reminiscent of fooling images (Fig.~\ref{fig:fooling}). Our attack also resembles deep dream\footnote{\url{https://en.wikipedia.org/wiki/DeepDream}}. Instead of altering the image to be classified in another class, we want it to be classified in the correct class, even with higher confidence. 









The formulation in Eq.~\ref{eq:def_revised} is basically a targeted attack in which the desired target class is actually the true class $y$. To increase the confidence of the model in choosing the label $y$ for image $x$, we can reduce the loss $J(x,y)$ by performing gradient descent on the image. Any existing attack can be reformulated to serve this purpose. Here, we use the iterative FGSM attack~\cite{kurakin2016physical,goodfellow2015explaining} (\aka IFGSM; commonly known as BIM) with $L_\infty$-norm:
\begin{eqnarray}
x'_{0} &=& x, \nonumber\\ 
x'_{N+1} &=& \text{clip}_{x, \epsilon}\Bigl\{ x'_{N} - \epsilon \sign \bigl( \nabla_xJ(x'_{N}, y)  \bigr) \Bigr\} 
\end{eqnarray}

    

\section{Experiments and results}




\begin{figure}[t]       
    \centering
    \includegraphics[width=\linewidth]{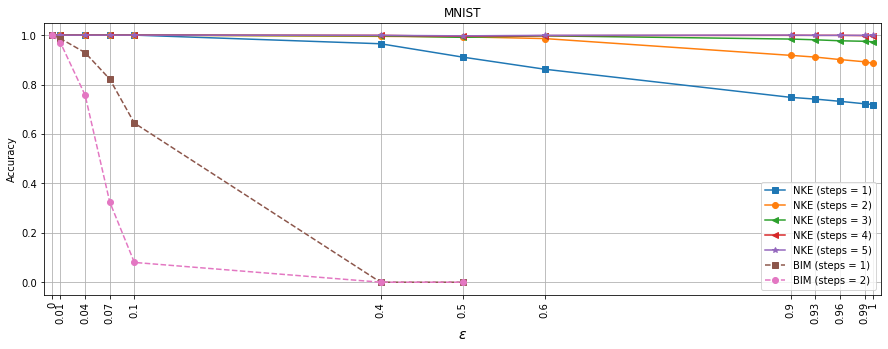}
    \includegraphics[width=\linewidth]{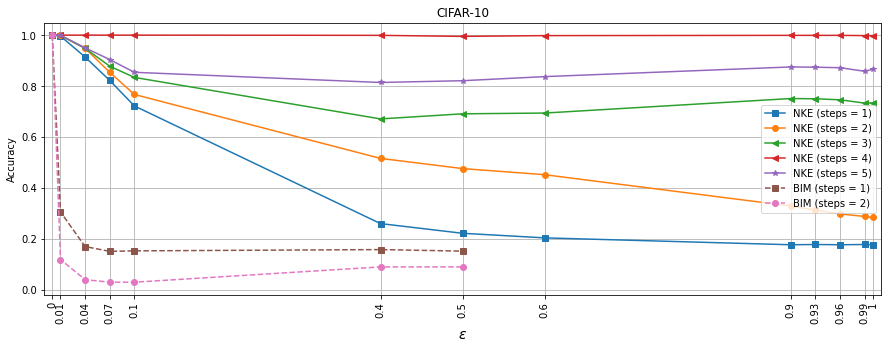}
    \caption{Attack performance. The lower accuracy means higher efficiency for the BIM attack. The opposite happens for the NKE attack. The $x$ axis shows the magnitude of perturbation. Please see also Fig.~\ref{fig:samplePreds} for sample images and their perturbed versions.}
    \label{fig:res}
\end{figure}

On MNIST, we train a CNN with two convolutional layers for 10 epochs. 
On CIFAR-10, we train a ResNet50 for 30 epochs. 
We do not use any data augmentation. Images are scaled to [0,1] range. The classifiers score about 95\% and 61\% test accuracy, respectively.

\begin{figure}[t]       
    \vspace{-15px}
    \centering
    \includegraphics[width=.3\linewidth]{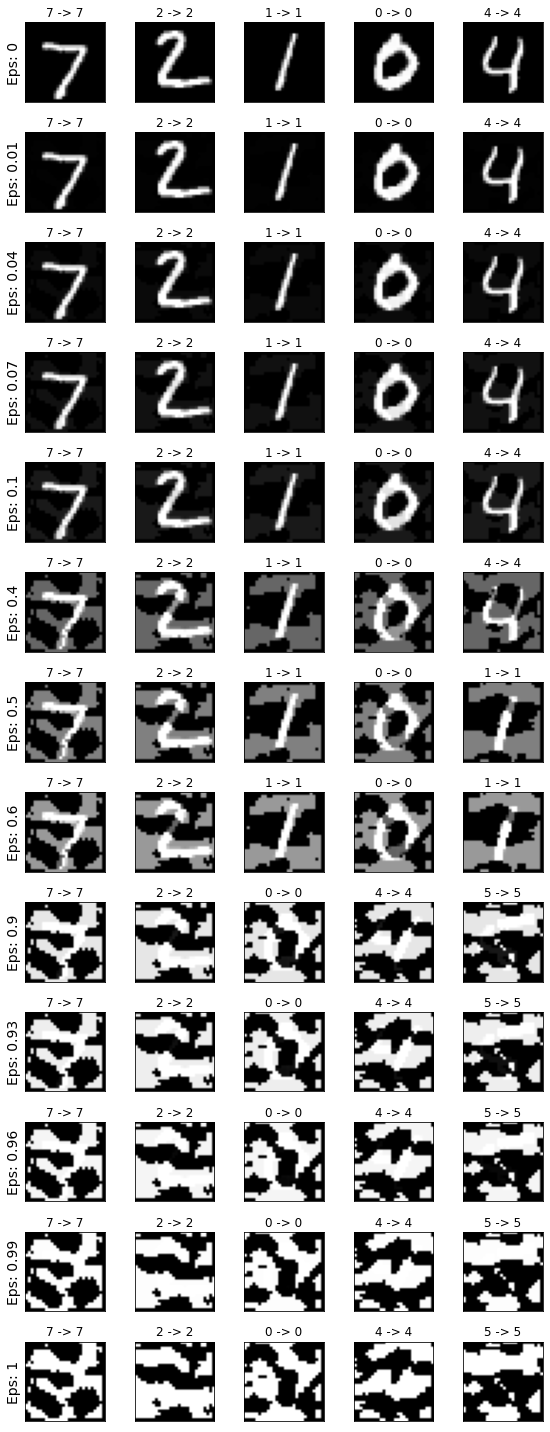} \hspace{13pt}
    \includegraphics[width=.3\linewidth]{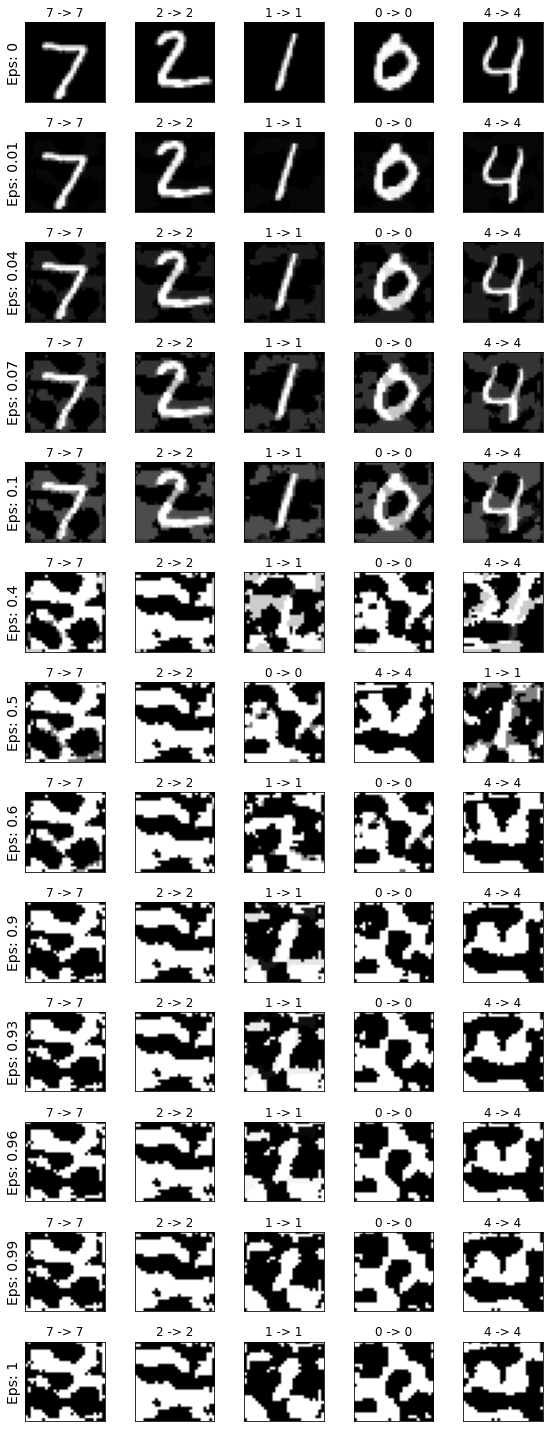} \hspace{13pt}
    \includegraphics[width=.3\linewidth]{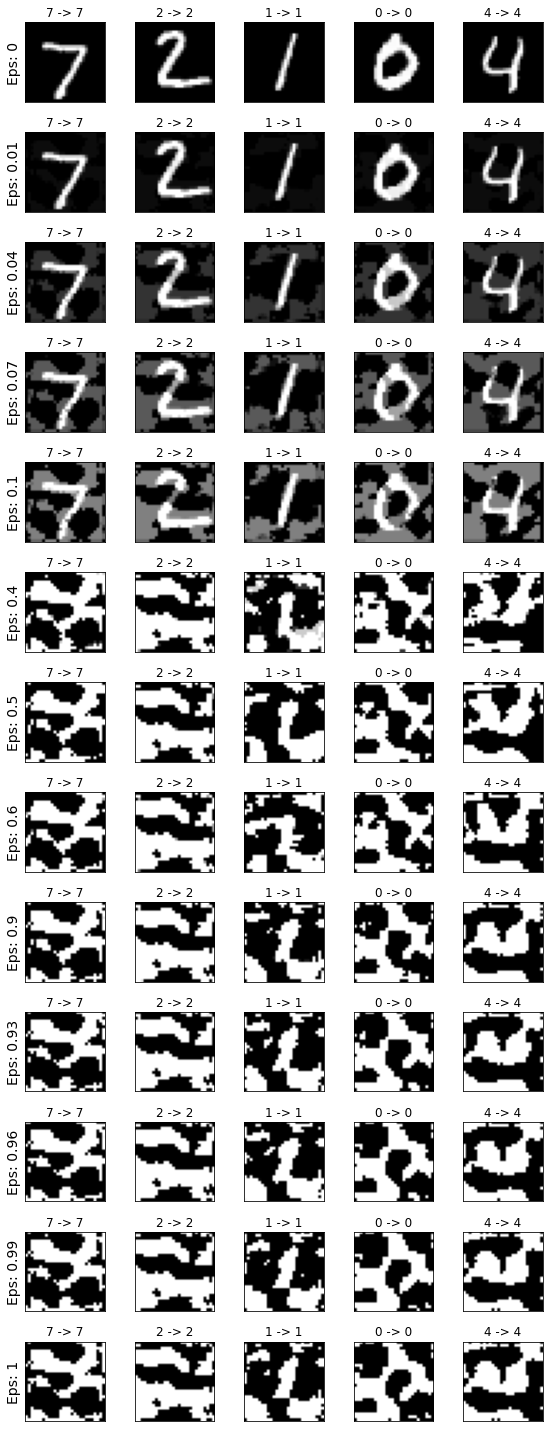} \\
    \includegraphics[width=.3\linewidth]{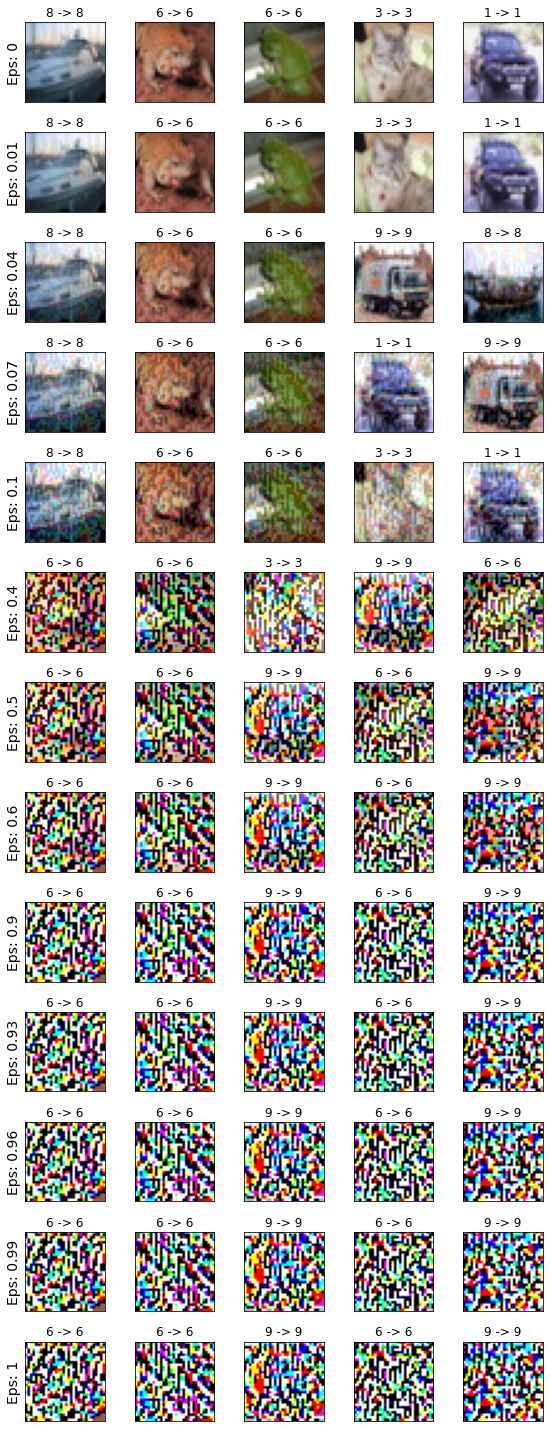} \hspace{13pt}
    \includegraphics[width=.3\linewidth]{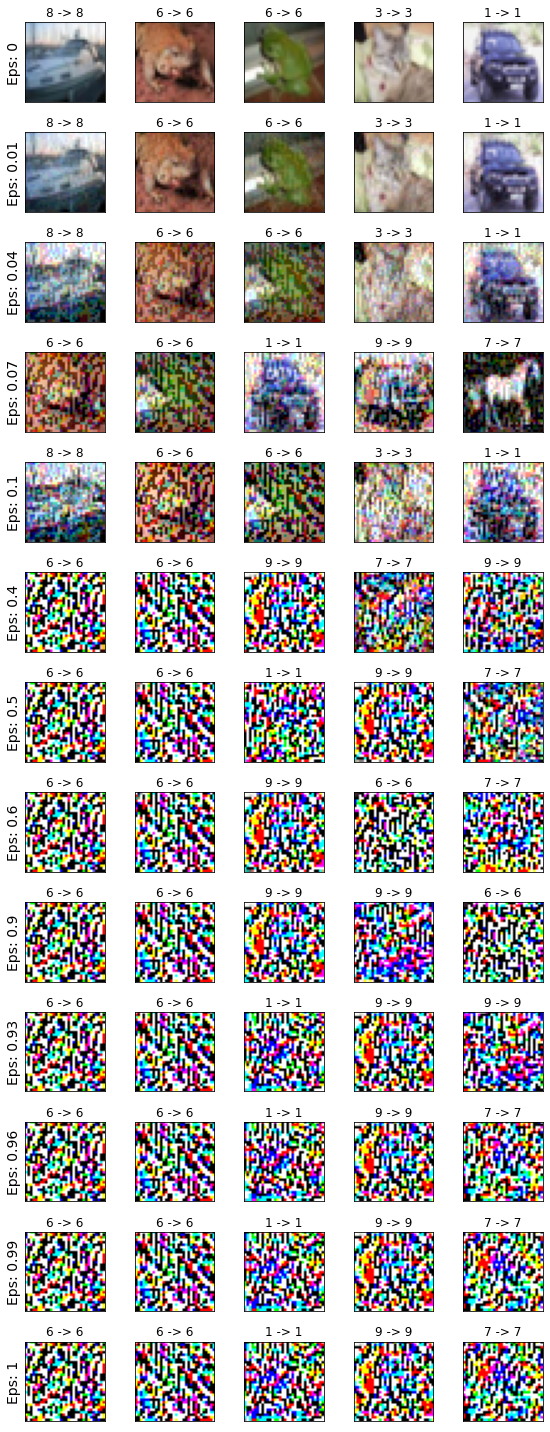} \hspace{13pt}
    \includegraphics[width=.3\linewidth]{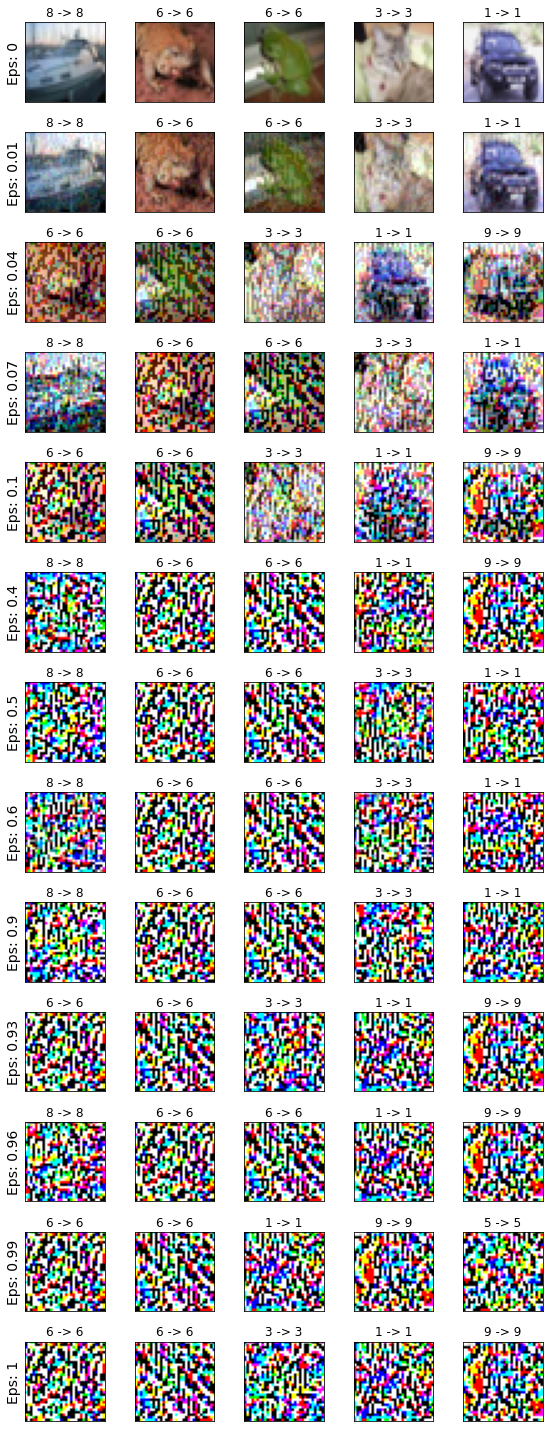}
    \vspace{-5px}
    \caption{Sample images from MNIST and CIFAR-10 datasets along with the adversarial examples generated by NKE attack for different values of $\epsilon$. Columns correspond to 1, 3, and 5 steps of NKE attack. Notice that $\epsilon_l$ drops as the number of steps grows which means that perturbations become visible sooner with more step of the attack.}
    \label{fig:samplePreds}
\end{figure}

To analyze the attack performance, we only consider the images that were classified correctly. We then measure the fraction of them that retain their labels after perturbation (\ie classified correctly again). Results are shown in Fig.~\ref{fig:res} and Fig.~\ref{fig:samplePreds}. The NKE attack is less effective as the perturbation grows. This is because there is more signal at small $\epsilon$ values, making it easy to alter the input to be classified correctly. Conversely, at higher perturbations when the signal fades, it becomes increasingly harder to alter the images such that its new label matchs the initial prediction. The NKE attack becomes more effective with more steps. With five steps, it is almost always possible to perturb the image such that it can be predicted correctly regardless of the magnitude of the signal in the image. Results are consistent across both datasets.

At $\epsilon=0.4$, the BIM attack with only one step completely punctures the MNIST classifier. Up to around $\epsilon_s=0.1$, perturbations generated by this attack are imperceptible. $\epsilon_s$ decreases with more steps of BIM. A similar observation can be made for the NKE attack. The image becomes almost completely unrecognizable at $\epsilon_l=0.9$ with one step of NKE. The classifier at this $\epsilon$ achieves around 75\% accuracy (\ie 75\% vs. 0\% by humans!). The NKE attack with five steps scores above 99\% at $\epsilon_l=0.9$ perturbation. Similar qualitative results are obtained on the CIFAR-10 dataset. Here, $\epsilon_l$ is about $0.4$ and $0.1$ corresponding to one and five steps of NKE, respectively.







\section{Conclusion and future work}
We introduced a new attack to generate a new kind of adversarial example. These adversarial images are similar to the fooling images but are much easier and faster to synthesize. Our work offers a new perspective in viewing adversarial attacks and unifies methods that generate adversarial examples and fooling images. Theoretically, any attack can be revised to generate {\bf adversarial examples that can fool humans, but not models}. Some may be more suitable than others for this purpose. For example, the Carlini-Wagner attack~\cite{carlini2017towards} can be revised as follows:
\begin{align}
    \text{Maximize} \ \  ||\epsilon||_2 \ \ s.t \\ 
        1. \ \ f(x+\epsilon) = f(x) \nonumber \\ 
        2. \ \  \ \ x + \epsilon  \in [0,1]^m \nonumber
\end{align}

which can be written as $ J(x + \epsilon, f(x))  - c \cdot ||\epsilon||_2  \ \ s.t \ \ x + \epsilon \in [0,1]^m$ using Lagrange multiplies, and can be optimized via the Quasi Newton method.

Some questions that should be answered in future research include: a) how transferable are the adversarial examples generated by NKE attack, b) is it possible to craft universal perceptible patterns that can make a classifier to maintain its initial decision for images from different classes (akin to~\cite{moosavi2017universal}), and c) how effective is adversarial training (or other defenses) against the NKE attack. Finally, we hope that our work will spark follow-up research and help gain deeper insights for solving the problem of adversarial vulnerability.






{\small
\bibliographystyle{plain}
\bibliography{refs}
}

\end{document}